\pdfoutput=1

\documentclass[11pt]{article}

\usepackage[]{acl}

\usepackage{times}
\usepackage{latexsym}

\usepackage[T1]{fontenc}

\usepackage[utf8]{inputenc}

\usepackage{microtype}
\usepackage{graphicx}
\usepackage{colortbl} 
\usepackage{array}    
\usepackage{listings}
\usepackage{xcolor}
\usepackage{multirow}
\usepackage{adjustbox}
\usepackage{makecell}
\usepackage{hhline}
\usepackage{booktabs}
\usepackage{enumitem}
\usepackage{soul} 
\usepackage{amsmath}
\usepackage{amsfonts}
\usepackage{caption, subcaption}
\usepackage[symbol]{footmisc}

\newcommand{\yz}[1]{}
\renewcommand{\yz}[1]{\textcolor{blue}{\textbf{Zhu}: #1}}
\newcommand{\ls}[1]{}
\renewcommand{\ls}[1]{\textcolor{magenta}{\textbf{LS}: #1}}
\newcommand{\lm}[1]{}
\renewcommand{\lm}[1]{\textcolor{red}{\textbf{LM}: #1}}

\title{Beyond Sparse Rewards: Enhancing Reinforcement Learning with Language Model Critique in Text Generation}

\author{{Meng Cao$^{1, 3, \dag, \ddag}$, Lei Shu$^{4}$, Lei Yu$^{2}$, Yun Zhu$^{4}$, Nevan Wichers$^{4}$, Yinxiao Liu$^{4}$, Lei Meng$^{4, \ddag}$}\\
    $^1$School of Computer Science, McGill University \\
    $^2$Department of Computer Science, University of Toronto \\
    $^3$Mila – Qu\'{e}bec AI Institute \\
    $^4$Google Research \\
}

\newcommand{\modelname}[0]{\text{RELC}}

\begin{document}
\maketitle

\def\thefootnote{\dag}\footnotetext{Work done during an internship at Google.}
\def\thefootnote{\ddag}\footnotetext{Correspondence to Meng Cao \texttt{<meng.cao@mail.mcgill.ca>} and Lei Meng \texttt{<leimeng@google.com>}}

\begin{abstract}
Reinforcement learning (RL) can align language models with non-differentiable reward signals, such as human preferences. 
However, a major challenge arises from the sparsity of these reward signals - typically, there is only a single reward for an entire output. This sparsity of rewards can lead to inefficient and unstable learning.
To address this challenge, our paper introduces an novel framework that utilizes the critique capability of Large Language Models (LLMs) to produce intermediate-step rewards during RL training.
Our method involves coupling a policy model with a critic language model, which is responsible for providing comprehensive feedback of each part of the output. This feedback is then translated into token or span-level rewards that can be used to guide the RL training process.
We investigate this approach under two different settings: one where the policy model is smaller and is paired with a more powerful critic model, and another where a single language model fulfills both roles.
We assess our approach on three text generation tasks: sentiment control, language model detoxification, and summarization. Experimental results show that incorporating artificial intrinsic rewards significantly improve both sample efficiency and the overall performance of the policy model, supported by both automatic and human evaluation.
\end{abstract}

\section{Introduction}
\begin{figure}[!t]
\setlength{\belowcaptionskip}{0.3cm}
\centering
\includegraphics[width=0.8\linewidth]{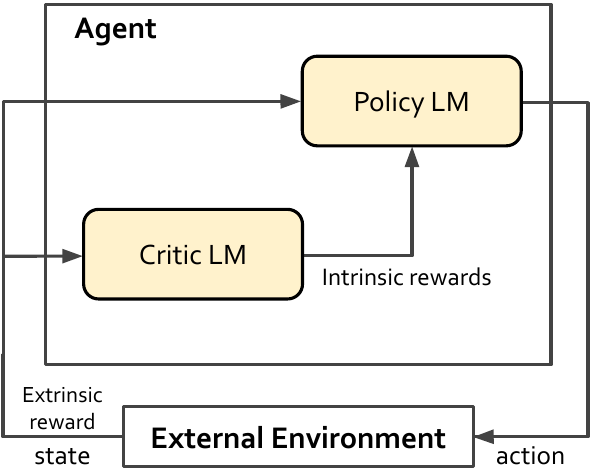}
\caption{Illustration of the proposed framework. There are two modules inside the agent. The critic LM takes the state and reward as input and generates dense intrinsic reward signals that evaluate different parts of the generation. The policy module is trained to optimize the weighted sum of intrinsic and extrinsic rewards.}
\label{fig:framework}
\end{figure}

Large language models (LLMs) have seen a rapid advancement in recent years, demonstrating a remarkable ability to understand and generate natural language \cite{brown2020language, touvron2023llama, OpenAI_GPT4_2023, biderman2023pythia, jiang2023mistral, shu2023rewritelm}. In the meanwhile, reinforcement learning has emerged as a complementary tool for further refining the capabilities of LMs. RL allows for the optimization of LMs towards any non-differentiable reward signal. For example, techniques like reinforcement learning from human feedback (RLHF) \cite{ziegler2019finetuning, NEURIPS2020_1f89885d} have been used to steer language models to better align with human preferences.

However, the reward signals received from the environment are usually sparse, a fundamental bottleneck that restricts the efficiency of learning \cite{NIPS2017_453fadbd, DBLP:conf/iclr/SukhbaatarLKSSF18}. Typically, in text generation tasks, a single scalar reward is obtained after a sentence or paragraph has been fully generated. This single reward signal introduces a temporal credit assignment problem, making it difficult for the model to learn which tokens were responsible for the received reward.
Previous attempts to circumvent the sparsity of rewards in RL have included reward shaping \cite{ng1999policy, devidze2022explorationguided, goyal2019using}, curiosity-driven exploration \cite{NIPS2016_afda3322, pathak2017curiosity, pmlr-v70-ostrovski17a}, and hierarchical RL \cite{NEURIPS2018_e6384711, zhang2021hierarchical}. However, these methods either require handcrafted features or do not translate straightforwardly into the domain of text generation.
A direct solution is to refine the environment's holistic reward model with one that offers dense rewards.
\citet{lightman2023let} and \citet{wu2023fine} have explored employing human annotators to provide detailed feedback at each intermediate step of model's generation. These annotations can then be used to train a fine-grained reward model. However, this method incurs high costs, and the resulting reward models tend to be highly task-specific, limiting their applicability across different tasks.

In light of these limitations, we introduce {\modelname} (\underline{Re}wards from \underline{L}anguage model \underline{C}ritique), an novel framework that leverages the critique capability of LLMs \cite{madaan2023self, saunders2206self, luo2023critique} to provide artificial reward signals for intermediate steps during RL training.
As illustrated in Figure~\ref{fig:framework}, we explicitly define an RL agent as the integration of 1) a policy model responsible for output generation, and 2) a critic model that tasked with assessing the quality of the outputs produced by the policy model. The critic LM, informed by the question, the policy model's output, and the single reward signal provided by the environment, generates verbal evaluation of each segment of the policy model's output. These evaluations are then converted into reward signals. We label the rewards generated by the critic model as ``intrinsic rewards'' to differentiate them from the reward signals provided by the environment. The critic model can be seamlessly integrated into RL algorithms such as PPO \cite{schulman2017proximal}, requiring no or very little modification to the algorithms themselves.

Our evaluation of the proposed method is carried out in two distinct settings: one employs a smaller policy model (GPT-2 Large) coupled with a more advanced critic model (GPT-3.5), and the other, a more challenging ``self-critique'' setting, where a single model (Llama 2) fulfills both roles.
We evaluate the effectiveness of our method through three text generation tasks: sentiment control, LM detoxification, and abstractive text summarization. 
The experimental results show that the use of LLM-generated intrinsic rewards significantly enhances sample efficiency across all tasks, with our approach outperforming established baseline methods according to both automated and human evaluation. Despite the additional inference cost incurred by incorporating the critic model, our approach is shown to be more computationally efficient, achieving superior performance to the baseline within the same computational budget.


\section{Related Work}
\paragraph{RL for Text Generation.} RL methods have been used in various text generation tasks including text summarization \cite{ryang-abekawa-2012-framework, pang2021text, dong-etal-2018-banditsum, cao-etal-2022-hallucinated}, machine translation \cite{NIPS2016_2f885d0f, DBLP:journals/corr/RanzatoCAZ15, NIPS2016_5b69b9cb, bahdanau2017an}, dialogue systems \cite{fatemi2016policy, li-etal-2016-deep, dhingra-etal-2017-towards, DBLP:journals/corr/abs-1907-00456} and question answering \cite{buck2018ask, xiong2018dcn, Nakano2021WebGPTBQ}. 
Recent studies have focused on combining RL with pre-trained language models like GPT-3 \citep{NEURIPS2020_1457c0d6} to generate text \citep{ouyang2022training, bai2022training, Nakano2021WebGPTBQ, NEURIPS2020_1f89885d} are better aligned with human preference such as being factual, relevant and helpful.

\paragraph{Reward Shaping and Intrinsic Rewards.}

\citet{ng1999policy} laid the groundwork for potential-based reward shaping in RL, demonstrating that such shaping can effectively reduce training time without changing the optimal policy.
\citet{NIPS2016_afda3322, pmlr-v70-ostrovski17a, NIPS2017_3a20f62a} have employed pseudo-count-based rewards to encourage exploration in environments where rewards are sparse.
\citet{NEURIPS2018_51de85dd} proposed a method where a parameterized intrinsic reward model is learned during training to generate dense reward signals.
This approach, however, presents certain optimization difficulties due to the necessity of calculating second-order gradients. 
\citet{wu2023fine, lightman2023let} employ human annotators to provide detailed span-level reward signals, demonstrating that these fine-grained rewards yield better performance compared to holistic rewards.

\paragraph{LLM for Reward Design.}
\citet{lee2023rlaif} employed an off-the-shelf LLM to create preference labels by comparing pairs of candidate responses. These labels were then used to train a holistic reward model.
Similarly, \citet{kwon2023reward} investigated the use of GPT-3 as an alternative to the actual reward function in RL training. Their method outperformed the reward model trained through supervised learning, yet it did not achieve the effectiveness of the true reward function.
\citet{du2023guiding, klissarov2023motif} use LLMs to generate rewards signals to encourage exploration of a gaming or robotic agent.
\citet{ma2023eureka} employed GPT-4 to generate the code for a reward function.

\section{Method}
The basic idea behind our method is to leverage LLM to generate dense intrinsic reward signal $r^{\text{in}}$ and provide it to an RL agent, which will optimize a combination of the intrinsic and extrinsic rewards. In this section, we first establish the Markov decision process (MDP) for text generation. Then, we discuss the policy gradient-based RL method widely used for text generation tasks. Finally, we detail the process of incorporating LLM-generated intrinsic rewards into RL training.

\begin{figure*}[!t]
\centering
\includegraphics[width=0.85\linewidth]{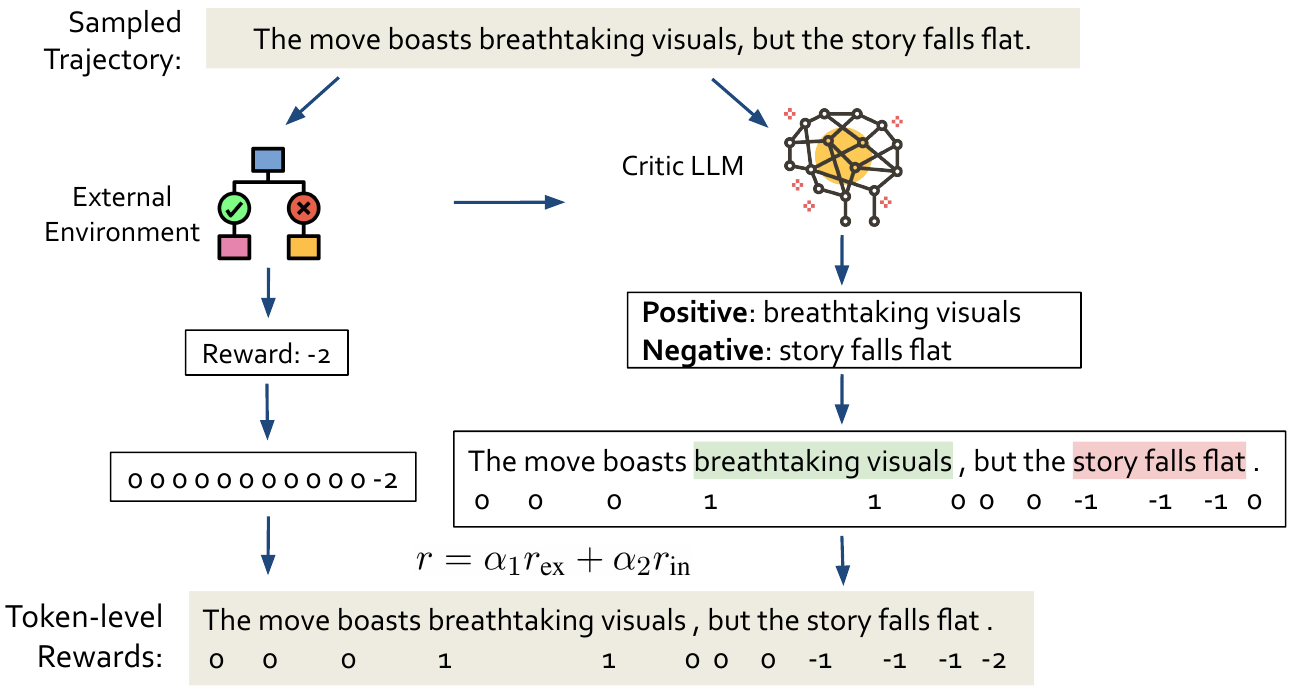}
\caption{An example demonstrating the reward calculation process in the sentiment control task. In this example, the external environment returns a scalar reward of -2 in response to the policy model's output. Subsequently, the critic model is prompted to identify spans of positive and negative sentiment within the output. Tokens within these spans are then assigned intrinsic rewards: +1 for positive and -1 for negative sentiment. The hyper-parameter $\alpha$ determines the weight of these two types of rewards. The extrinsic reward is assigned to the last position in the output sequence.}
\label{fig:sent_process}
\end{figure*}

\subsection{RL for Text Generation}
Let us consider the language generation procedure as a MDP \citep{puterman1994markov}, defined by the tuple $(\mathcal{S}, \mathcal{A}, \mathcal{P}, \mathcal{R}, \gamma)$. Here, $\mathcal{S}$ represents the set of all possible states, \(\mathcal{A}\) is the set of actions, $\mathcal{P}: \mathcal{S}\times\mathcal{A}\times\mathcal{S}\mapsto [0,1]$ is the state transition function, $R: \mathcal{S}\times\mathcal{A}\times\mathcal{S}\mapsto \mathbb{R}$ is the reward function assigning a numerical value to each transition $(s, a, s')$, and $\gamma \in [0,1]$ is the discount factor. In the context of text generation, we operate under the assumption of an episodic, discrete-actions, RL setting.
The input prompt $s_0 \in \mathcal{S}$ sets the starting state. At each decoding step $t$, the state $s_t\in\mathcal{S}$ consists of the prompt and the concatenation of the previously generated tokens. Choosing an action involves selecting a token from the vocabulary, leading to a new state \(s_{t+1}\), created by appending the selected token to the currently generated partial sentence. 
The agent's policy $\pi_{\theta}(a|s)$, which is a language model parameterized by $\theta$, determines the probability of selecting each action at a given state. 
The goal of the agent is to maximize the discounted cumulative reward throughout the trajectory: $J(\theta) = \mathbb{E}_{\tau \sim \pi_{\theta}} \Big[ \sum_{t=0}^T \gamma^t r_t \Big]$.

\subsection{Policy Gradient based RL \& PPO}
Policy gradient methods, which are commonly applied in text generation, directly parameterize the policy model to optimize its parameters $\theta$ with the goal of maximizing $J(\theta)$. The gradient $\nabla_{\theta} J(\theta)$ is proportional to the expectation of the product of the gradient of the log policy and the return $G_t$ \cite{NIPS1999_464d828b}:
\begin{equation}\label{eq:policy_gradient}
    \nabla_{\theta} J(\theta) = \mathbb{E}_{\pi_{\theta}}\left[\nabla_{\theta} \log \pi_{\theta}(a_t|s_t) G_t\right] 
\end{equation}
where the return is defined as $G_t=\sum_{i=t}^T\gamma^{i-t}r_i$. A high return leads to the reinforcement of all actions by increasing their selection probability. To reduce variance, a widely adopted strategy involves substituting the raw return $G_t$ in Equation~\ref{eq:policy_gradient} with a generalized advantage estimation function \cite{Schulmanetal_ICLR2016}: $$\hat{A}_t=\sum^T_{t'=t}(\gamma\lambda)^{t'-t}(r_{t'}+\gamma V(s_{t'+1})-V(s_{t'})))$$ where $\lambda$ is a hyper-parameter and $V(s_{t'})$ is the value function representing the expected return at state $s_{t'}$.
Several variants of the basic policy gradient approach have been proposed to improve training stability. One widely used variant, particularly in the context of text generation, is Proximal Policy Optimization (PPO) \cite{schulman2017proximal}. PPO introduces mechanisms to stabilize the training process by limiting the updates to the policy at each step, effectively preventing destructive large updates that can cause the policy to perform worse. In this work, we use the clipped surrogate objective function of PPO which is expressed as:
\begin{equation*}
\resizebox{\columnwidth}{!}{
$L(\theta)=\hat{\mathbb{E}}_{t} \Big[ \min(r_t(\theta) \hat{A}_t, 
\text{clip}(r_t(\theta), 1-\epsilon, 1+\epsilon) \hat{A}_t) \Big]$
}
\end{equation*}
where $r_t(\theta)=\frac{\pi_\theta(a_t|s_t)}{\pi_{\theta_{\text{old}}}(a_t|s_t)}$ is the probability of taking action $a_t$ at state $s_t$ in the current policy divided by the previous one.

\subsection{Learning with LLM Generated Intrinsic Rewards}
The current RL frameworks for text generation, such as RLHF \cite{ziegler2019finetuning, NEURIPS2020_1f89885d}, the environment takes the entire generated text as input and returns a scalar score. Therefore, the learning typically depends on a sparse reward signal that becomes accessible only upon the generation of a complete sentence. 
We refer to this reward signal as the extrinsic reward $r^\text{ex}$ and we have $r^\text{ex}_{t<T}=0$.
Our method deviates from the existing approaches by differentiating between the extrinsic reward from the environment and an additional intrinsic reward $r^{\text{in}}$ generated by LLM. 
As shown in Figure~\ref{fig:framework}, within the agent, our framework incorporate an additional critic language model alongside the policy model. The task of the critic model is to pinpoint the tokens or segments in the policy's output that directly contribute to receiving the environment's reward. The critic model is fed with a task description $D$, a set of few-shot examples $E$, the current state $s$ as determined by the policy model's output, and optionally, the reward $r^\text{ex}$ received from the environment. 
For token at step $t$, if it is part of the identified segment, we assign an non-zero value to the intrinsic reward $r_t^\text{in}$.
The final reward is defined as the weighted sum of extrinsic and intrinsic rewards: $r(s,a)=\alpha_1 r^\text{ex}(s,a) + \alpha_2 r^\text{in}(s,a)$ where $\alpha_1$ and $\alpha_2$ are hyper-parameters that controls the weight of the reward. Note that extrinsic rewards are only non-zero at the final time step, specifically when $t=T$. The policy LM is optimized to maximize the combined reward: $J(\theta)^{{\modelname}} = \mathbb{E}_{\tau \sim \pi_{\theta}} \Big[ \sum_{t=0}^T \gamma^t (\alpha_1 r^\text{ex} + \alpha_2 r^\text{in}) \Big]$ where the policy model is parameterized by $\theta$. The critic LM is frozen during training. In this work, we employed the PPO algorithm to train the agent. However, it's worth noting that our framework is versatile and can also be integrated with other reinforcement learning algorithms, such as Advantage Actor-Critic (A2C) \cite{pmlr-v48-mniha16}. An illustration of how rewards are calculated in the sentiment control task is provided in Figure~\ref{fig:sent_process}. In subsequent sections, references to PPO specifically denote the use of PPO with extrinsic rewards only.

\paragraph{LLM Choice and Prompt Design} In this work, we employ two LLMs as critics: \texttt{gpt-3.5-turbo} and 7B Llama2 \cite{touvron2023llama}. The input prompt is structured in three segments. First, we define the task within the prompt, outlining the types of correct responses or errors the critic model should identify. For instance, in the detoxification task, we clearly specify what constitutes toxic language.
Next, we include a curated set of few-shot examples (3-shot unless otherwise specified). These examples are chosen meticulously to include a broad spectrum of exemplary responses and typical errors produced by the policy model. Finally, we give the critic model the current question, the output from the policy model, and, optionally, the extrinsic reward from the environment. This extrinsic reward is incorporated to better align the critic's evaluation with the desired outcomes. It's important to note that our primary goal is to optimize the agent towards extrinsic reward, as these are the ultimate indicators of performance. Intrinsic rewards, on the other hand, are used only for providing immediate feedback and enhancing the learning process. As such, we want to ensure that the critic models' feedback and the extrinsic rewards are well aligned.
The specifics of the prompt used are detailed in the Appendix.

\section{Experiments}
\label{sec:experiment}
In this section, we demonstrate that our method outperforms the PPO baseline in three text generation tasks: sentiment control, LM detoxification, and text summarization.

\subsection{Sentiment Control}\label{subsec:sentiment}
In the sentiment control task, the objective is to guide the LM towards producing responses with a positive sentiment, starting from prompts that are neutral or negative.

\subsubsection{Experimental Setup}
We consider two settings: 1) a small policy LM (GPT-2 large) paired with a strong critic LM (\texttt{gpt-3.5-turbo}); 2) the policy model and critic model are the same, which we use Llama 2 \cite{touvron2023llama} as initialization.
We access the \texttt{gpt-3.5-turbo} model through OpenAI's API.
For training, we make use of the IMDB dataset that contains 25K movie reviews \cite{maas-EtAl:2011:ACL-HLT2011}. We randomly extract the first 4 to 10 tokens from each review as the input prompt. The policy model is trained on the training set for one epoch. We set $\alpha_1=1$ and $\alpha_2=0.2$.
Following the experimental setup of \cite{liu-etal-2021-dexperts, Lu2022QuarkCT}, we use the OpenWebText (OWT) Corpus dataset \citep{Gokalsan2019OpenWeb} as our test set. \citet{liu-etal-2021-dexperts} curated three distinct test sets from OWT: \textit{neutral} (5K prompts), \textit{positive} (2.5K prompts), and \textit{negative} (2.5K prompts). These sets were created based on the likelihood of the prompt leading to positive or negative continuations.
For the reward model, we employ a distilled BERT classifier that is trained on the IMDB dataset\footnote{\url{https://huggingface.co/lvwerra/distilbert-imdb}}. Details regarding the prompts, few-shot examples, and additional hyper-parameters can be found in Appendix~\ref{appendix:sent}.


\paragraph{Baselines and evaluation metrics}

We compare our method with seven baseline methods including PPLM \citep{Dathathri2020Plug}, CTRL \citep{keskar2019ctrl}, DAPT \citep{gururangan-etal-2020-dont}, GeDi \cite{krause-etal-2021-gedi-generative}, \textsc{DExperts} \citep{liu-etal-2021-dexperts}, \textsc{Rect} \citep{cao2023systematic}, and PPO.
For sentiment evaluation, we adopt the approach of \citet{liu-etal-2021-dexperts, Lu2022QuarkCT} and calculate the average percentage of positive/negative continuations from the 25 generated outputs using HuggingFace's sentiment analysis classifier fine-tuned on SST-2. Moreover, we analyze fluency and diversity to measure how each method impacts the overall text quality. We use GPT-2 XL perplexity (PPL) as a proxy for fluency. For diversity, we calculate the normalized count of unique bigrams.

\subsubsection{Results}


\begin{figure}[!t]
\centering
\begin{minipage}{\linewidth}
    \centering\captionsetup[subfigure]{justification=centering}
    \includegraphics[width=0.82\linewidth]{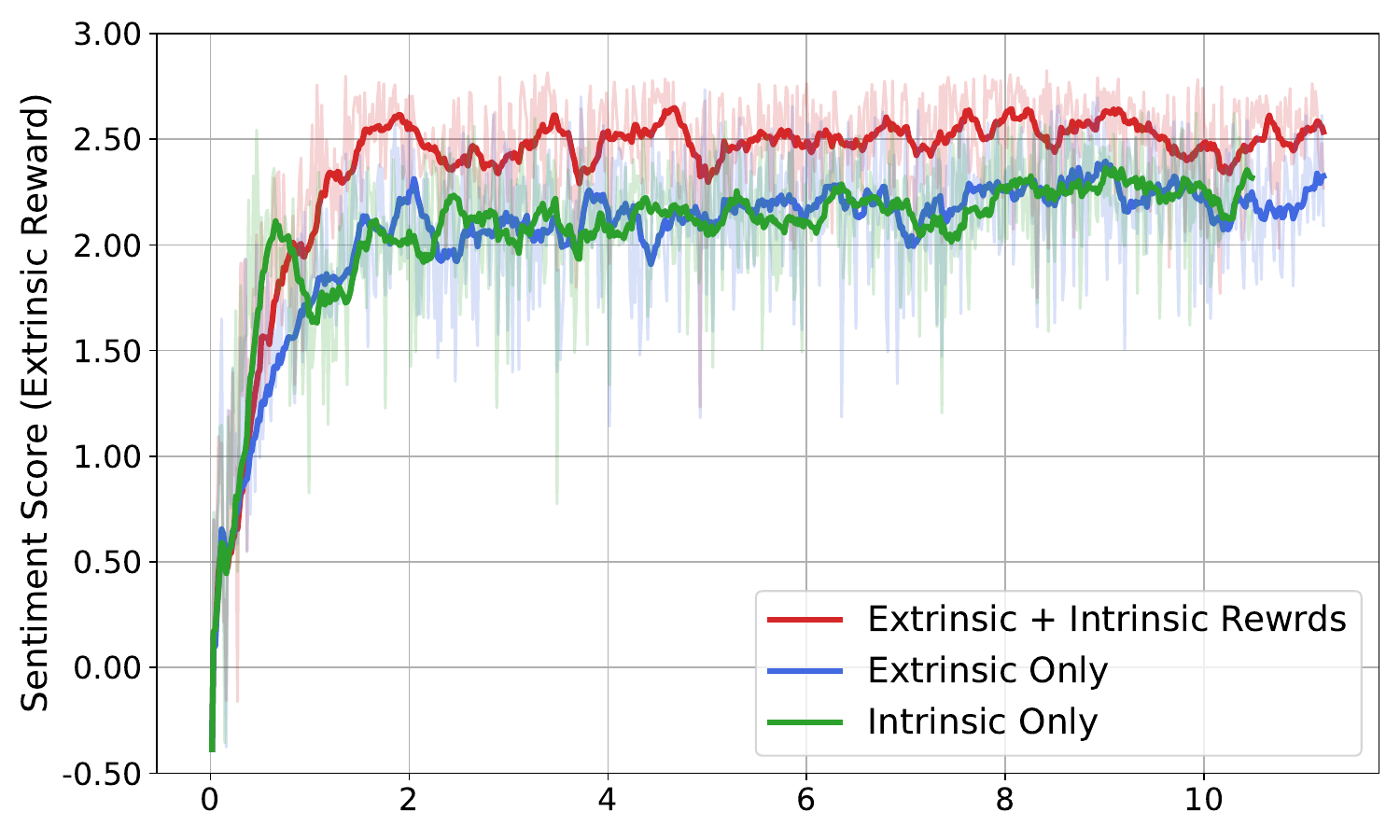}
    \subcaption{GPT-2 large as policy LM and GPT-3.5 as critic}
    \label{subfig:sent_gpt2}\par\vfill
    \includegraphics[width=0.82\linewidth]{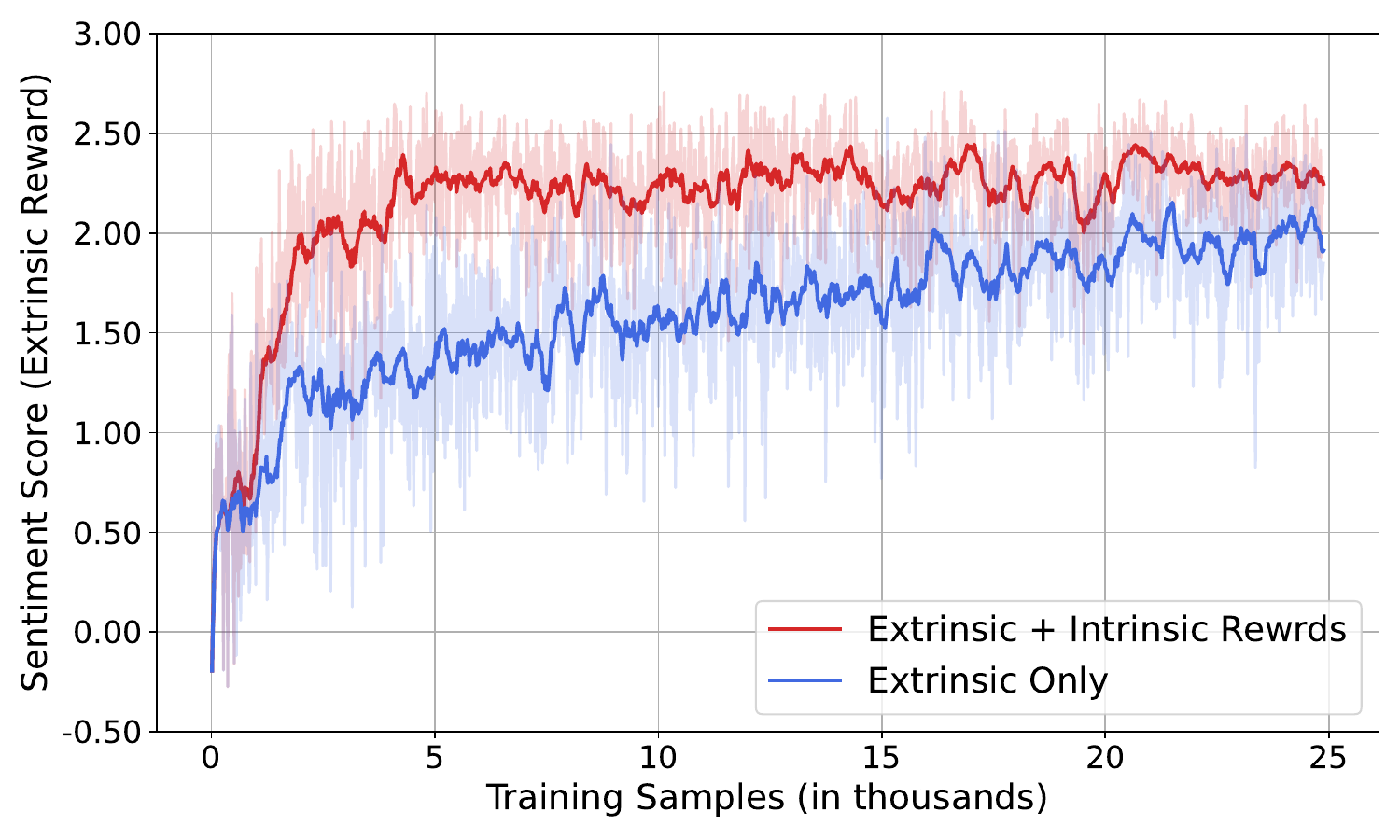}
    \subcaption{Self-critique using Llama 2 7B}
    \label{subfig:sent_llama2}
\end{minipage}
\caption{\label{fig:learning_curve_sent} Learning curves of the sentiment control experiment on the IMDB dataset. The x-axis is the number of training samples, while the y-axis shows the extrinsic reward, defined as the logit of the positive class returned by a distilled BERT sentiment classifier. The curves are smoothed using a moving average of 10 to improve readability.}
\vskip -0.1in
\end{figure}

\begin{table}
\renewcommand{\arraystretch}{0.95}
\centering
\small
\vspace{-3mm}
\begin{tabular}{lccccc}
\toprule
~ & \multicolumn{2}{c}{\bf \% Positive ($\uparrow$)} & {\bf Fluency} & \multirow{2}{*}{\bf Dist. ($\uparrow$)} \\
~ & neg. & neu. & ppl. ($\downarrow$) & ~ \\
\midrule
GPT2 (large) & 0.00 & 50.02 & 11.31 & 0.85 \\
\midrule
PPLM & 8.72 & 52.68 & 142.1 & \bf 0.86 \\
CTRL & 18.88 & 61.81 & 43.79 & 0.83 \\
GeDi & 26.80 & 86.01 & 58.41 & 0.80 \\
\textsc{DExperts} & 36.42 & 94.46 & 25.83 & 0.84 \\
DAPT & 14.17 & 77.24 & 30.52 & 0.83 \\
PPO & 43.13 & 94.10 & 15.16 & 0.80 \\
\textsc{Quark} & 46.55 & 95.00 & 14.54 & 0.80 \\
\midrule
{\modelname} & \bf 59.06 & \bf 95.63 & \bf 13.79 & 0.80 \\
\bottomrule
\end{tabular}
\caption{\label{tab:sentiment} Automatic evaluation results of the sentiment control experiments. All models are based on GPT2-large. Baseline results are reported in \citet{liu-etal-2021-dexperts, Lu2022QuarkCT}. \textbf{Neg.} column shows the evaluation results on 2.5K negative prompts and \textbf{Neu.} shows the evaluation results on 5K neutral prompts.}
\end{table}

Figure~\ref{fig:learning_curve_sent} presents the learning curves for both our method and baselines. From the figure, we can find that {\modelname} has better sample efficiency compared to the baselines in both settings. Table~\ref{tab:sentiment} shows the evaluation results on the OWT Corpus test set. As shown in the table, our method outperforms all the baselines in terms of steering towards positive sentiment. Besides, compared to the baseline, our model has the least impact on the fluency of generated sentences.

\subsection{LM Detoxification}
\label{subsec:detoxificatoin}
In this experiment, we focus on the task of LM detoxification. We show that the integration of LLM-generated intrinsic rewards into RL training can improve both sample efficiency and the final detoxification performance.

\subsubsection{Experimental Setup}
In our detoxification experiments, we utilize the \textsc{RealToxicityPrompts (RTP)} benchmark \citep{gehman-etal-2020-realtoxicityprompts} for training and evaluation. RTP contains 100K human-written sentence prefixes (i.e., prompts) derived from English web texts.
For toxicity evaluation, we utilize the Perspective API\footnote{\url{https://github.com/conversationai/perspectiveapi}}, a tool that is widely used in previous work for automatic toxicity evaluation. The API provides a score from 0 to 1, with 1 indicating high toxicity and 0 signifying non-toxic content.
Following the experimental setup of previous work, we use 85K of these prompts as training set. Our evaluation is conducted on the 10K non-toxic test prompts used by \citet{liu-etal-2021-dexperts, Lu2022QuarkCT}.
We use $1-\textsc{Perspective}\,(y)$ as the reward signal. We set $\alpha_1$ to 1 and $\alpha_2$ to 0.5.
More information about the prompts, few-shot examples, and hyper-parameters can be found in Appendix~\ref{appendix:tox}.

\paragraph{Baselines and evaluation metrics.}
We conducted a comparative analysis of our method against seven baseline methods. Out of these, six are the same as those discussed in Section~\ref{subsec:sentiment}. Additionally, we add another baseline method \textsc{Rect}  \cite{cao2023systematic}.
We report two metrics: the average of maximum toxicity scores over 25 generations and the empirical probability of a toxic continuation appearing at least once over 25 generations.

\subsubsection{Results}
\begin{figure}[!t]
\centering
\begin{minipage}{\linewidth}
    \centering\captionsetup[subfigure]{justification=centering}
    \includegraphics[width=0.8\linewidth]{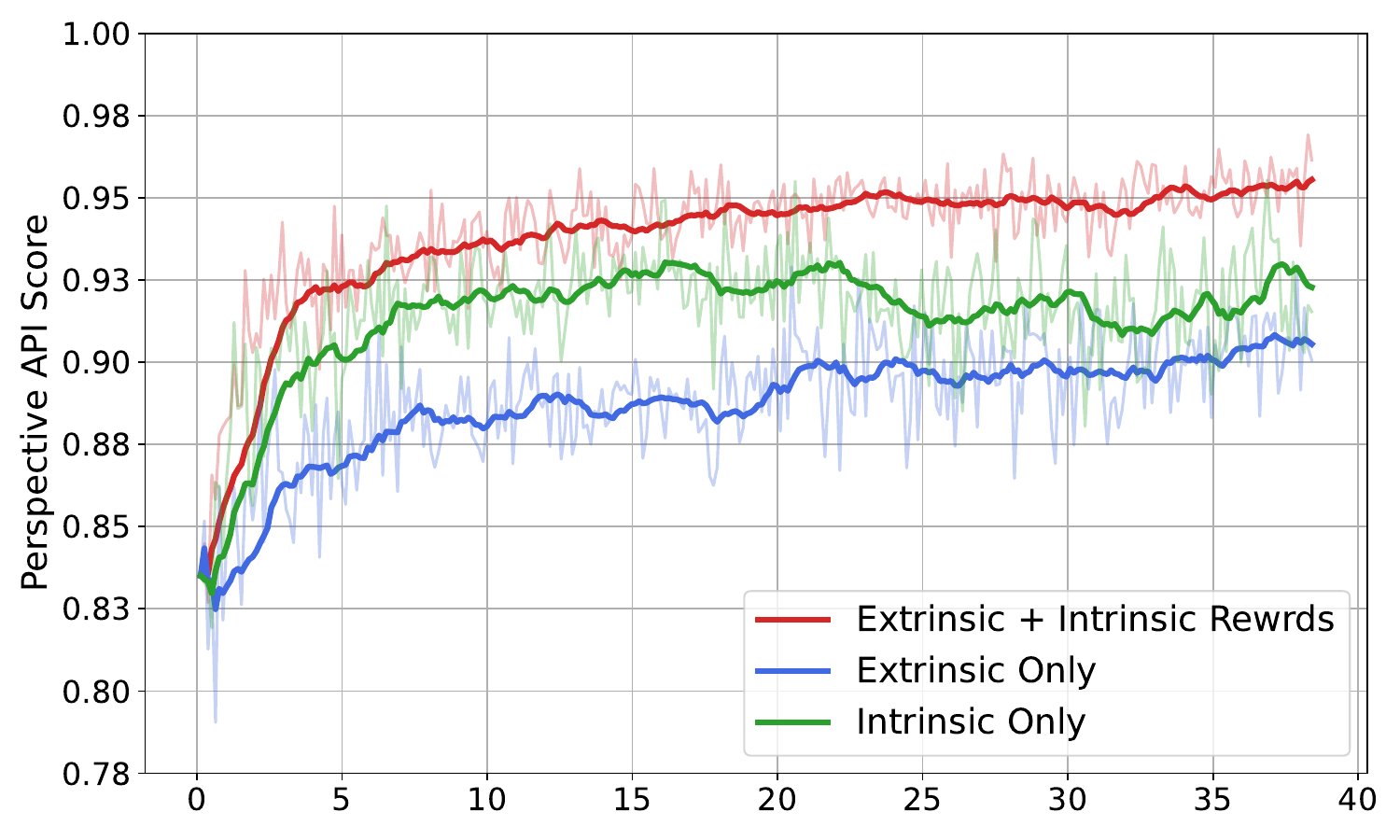}
    \subcaption{GPT-2 large as policy LM and GPT-3.5 as critic}
    \label{subfig:tox_gpt2}\par\vfill
    \includegraphics[width=0.8\linewidth]{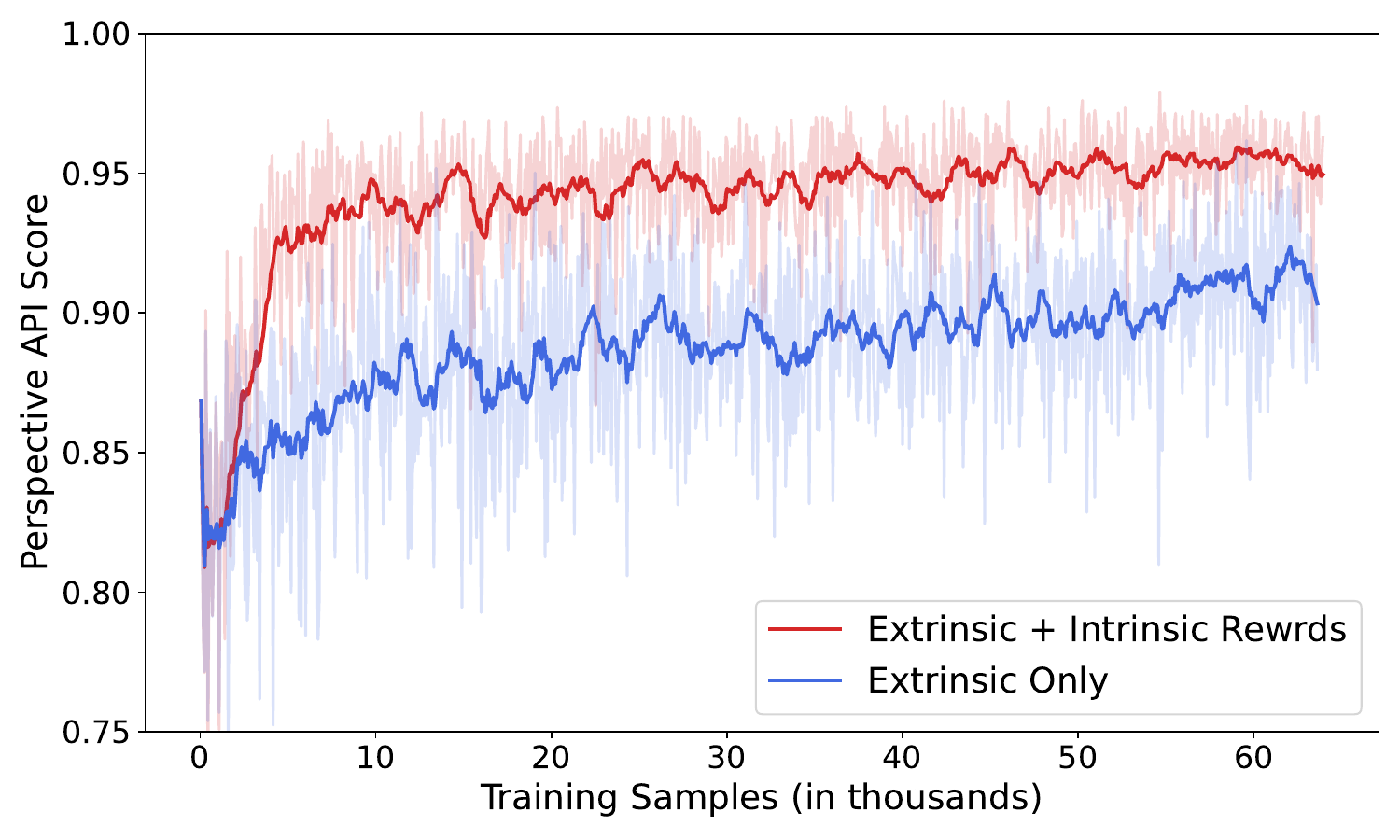}
    \subcaption{Self-critique using Llama 2 7B}
    \label{subfig:tox_llama2}
\end{minipage}
\caption{\label{fig:learning_curve_tox} Plot shows the learning curves of the detoxification experiment on the RTP dataset, smoothed using a moving average of 10 to improve readability. X-axis shows the number of training samples (in thousands) and y-axis is the average of non-toxic probability (same as extrinsic reward) measured using Perspective API.}
\vskip -0.1in
\end{figure}

\begin{table}
\renewcommand{\arraystretch}{0.95}
\centering
\small
\vspace{-3mm}
\begin{tabular}{lccccc}
\toprule
~ & \multicolumn{2}{c}{\bf Toxicity ($\downarrow$)} & {\bf Fluency} & \multirow{2}{*}{\bf Dist. ($\uparrow$)} \\
~ & avg.max. & \%prob. & ppl. ($\downarrow$) & ~ \\
\midrule
GPT2 & 0.527 & 52.0 & 11.31 & 0.85 \\
\midrule
PPLM & 0.520 & 51.8 & 32.58 & \bf 0.86 \\
GeDi & 0.363 & 21.7 & 60.03 & 0.84 \\
\textsc{DExperts} & 0.314 & 12.8 & 32.14 & 0.84 \\
DAPT & 0.428 & 36.0 & 31.21 & 0.84 \\
Rect & 0.266 & 7.9 & - & \bf 0.86 \\
PPO & 0.218 & 4.4 & 14.27 & 0.80 \\
\textsc{Quark} & 0.196 & 3.5 & 12.47 & 0.80 \\
\midrule
{\modelname} & \bf 0.133 & \bf 0.7 & \bf 11.72 & 0.80 \\
\bottomrule
\end{tabular}
\caption{\label{tab:detoxification}
Detoxification evaluation results on 10K non-toxic prompts from the \textsc{RealToxicityPrompts} dataset, using the identical test set as referenced in \citet{gehman-etal-2020-realtoxicityprompts, liu-etal-2021-dexperts}.
We use top-$p$ sampling with $p=0.9$ to sample up to 20 tokens.
Baseline results are from \citet{Lu2022QuarkCT} and \citet{cao2023systematic}.
}
\end{table}

\begin{table}[h]
\renewcommand{\arraystretch}{0.95}
\centering
\small
\begin{tabular}{lcccc}
\toprule
~ & {\bf Toxicity ($\downarrow$)} & {\bf Fluency} & {\bf Dist. } \\
~ & avg.max. & ppl. ($\downarrow$) & dist-3 ($\uparrow$) \\
\midrule
{F.G. RLHF} & 0.081 & 9.77 & 0.932 \\
{\modelname} & \bf 0.050 & \bf 9.53 & \bf 0.934 \\
\bottomrule
\end{tabular}
\caption{\label{tab:tox_fgrlhf_compare} Comparison with Fine-Grained RLHF \cite{wu2023fine}. In alignment with the experimental setting of \citet{wu2023fine}, we use nucleus sampling decoding with $p=0.9$ and temperature $=1.0$. The generation length limit is set to 48.
}
\end{table}
As shown in Figure~\ref{fig:learning_curve_tox}, incorporating intrinsic rewards greatly improves sample efficiency. Another interesting find is that using only intrinsic rewards also outperforms extrinsic reward baseline. Table~\ref{tab:detoxification} shows the evaluation results on the test set. As shown in the table, our method significantly reduces the rate of toxic generations compared to all baseline methods. Moreover, our approach has a minimal effect on fluency, as measured by perplexity, while also maintaining a similar level of diversity. In Table~\ref{tab:tox_fgrlhf_compare}, we compare our method with Fine-Grained RLHF \cite{wu2023fine} which queries the API using partial generated sentences to get fine-grained rewards. As shown in Table~\ref{tab:tox_fgrlhf_compare}, our method outperforms Fine-Grained RLHF in terms of both detoxification performance and text quality. 
We also directly prompt Llama 2 with detoxification instructions and few-shot examples. As shown in Table~\ref{tab:tox_llama2_prompting}, our method outperforms the prompting-based method.

\begin{table}
\renewcommand{\arraystretch}{0.95}
\centering
\small
\begin{tabular}{lcccc}
\toprule
~ & \multicolumn{2}{c}{\bf Toxicity ($\downarrow$)} & \multirow{2}{*}{\bf Dist. ($\uparrow$)} \\
~ & avg.max. & \%prob. & ~ \\
\midrule
Llama 2 & ~ & ~ & ~ \\
\:+ 1-shot & 0.409 & 30.9 & 0.77 \\
\:+ 3-shot & 0.426 & 32.5 & 0.78 \\
\midrule
PPO & 0.276 & 12.53 & 0.82 \\
{\modelname} & 0.176 & 3.68 & 0.82 \\
\bottomrule
\end{tabular}
\caption{\label{tab:tox_llama2_prompting} Llama2 evaluation results for the detoxification task. Prompt and few-shot examples used can be found at Appendix~\ref{appendix:tox}.}
\end{table}

\subsection{Summarization}
In this section, we demonstrate how our approach effectively improves the language model's ability to generate summaries that are better aligned with human preference.

\subsubsection{Experimental Setup}
We use the Reddit TL;DR dataset \cite{volske-etal-2017-tl} for the summarization experiment. The dataset contains approximately 3 million posts gathered from \texttt{reddit.com}, spanning a wide range of topics. We employ the filter version of the original dataset as provided by \citet{NEURIPS2020_1f89885d}, which consists of around 116K training samples, 6K validation and test samples. 
We fine-tuned a GPT2-large model via supervised learning on the whole training set for 9,000 steps, using a batch size of 64. This model serves as the initialization for the policy model. For RL training, we fine-tuned the policy model on 30K training samples for one epoch.
Following \citet{NEURIPS2020_1f89885d}, our reward model is a 6B language model fine-tuned on 92K human-annotated pairwise summary comparison dataset. The reward model achieves 75.94\% accuracy on the validation set. We set $\alpha_1=1.0$ and $\alpha_2=0.1$.
We use \texttt{gpt-3.5-turbo} for generating intrinsic rewards in a 3-shot setting. Details regarding the prompt, the few-shot examples, and the hyper-parameters applied in the summarization experiment are provided in Appendix~\ref{appendix:summ}.

\paragraph{Baselines and evaluation metrics.} We compare our method with two baseline methods: the supervised fine-tuning baseline (SFT) and the PPO baseline.
For summary quality evaluation, we use both the ROUGE score and the preference score calculated using the reward model. It worth mentioning that ROUGE score is not often reliable and doesn't capture human preference. As shown in \citet{NEURIPS2020_1f89885d}, the preference score consistently outperforms the ROUGE score with a better agreement with the human annotators on summary quality.

\subsubsection{Results}
\begin{table}
\renewcommand{\arraystretch}{0.95}
\centering
\small
\vspace{-3mm}
\begin{tabular}{lccc}
\toprule
~ & \multicolumn{2}{c}{\bf Rouge ($\uparrow$)} & \multirow{2}{*}{\bf Pref. Score ($\uparrow$)} \\
~ & R-1 & R-L & ~ \\
\midrule
SFT & 34.78 & 26.97 & 2.34 \\
PPO & 30.81 & 22.11 & 3.25 \\
{\modelname} & 29.32 & 20.17 & \bf 3.88 \\
\bottomrule
\end{tabular}
\caption{\label{tab:summ} Summarization task evaluation results on the Reddit TL;DR test set, with \textbf{Pref. Score} representing the preference score calculated using a GPT-J-6B model \cite{gpt-j} fine-tuned on a human preference dataset \cite{NEURIPS2020_1f89885d}.
}
\end{table}


\begin{figure}[!t]
\centering
\captionsetup{font=small}
\begin{minipage}{\linewidth}
    \centering\captionsetup[subfigure]{justification=centering}
    \includegraphics[width=0.8\linewidth]{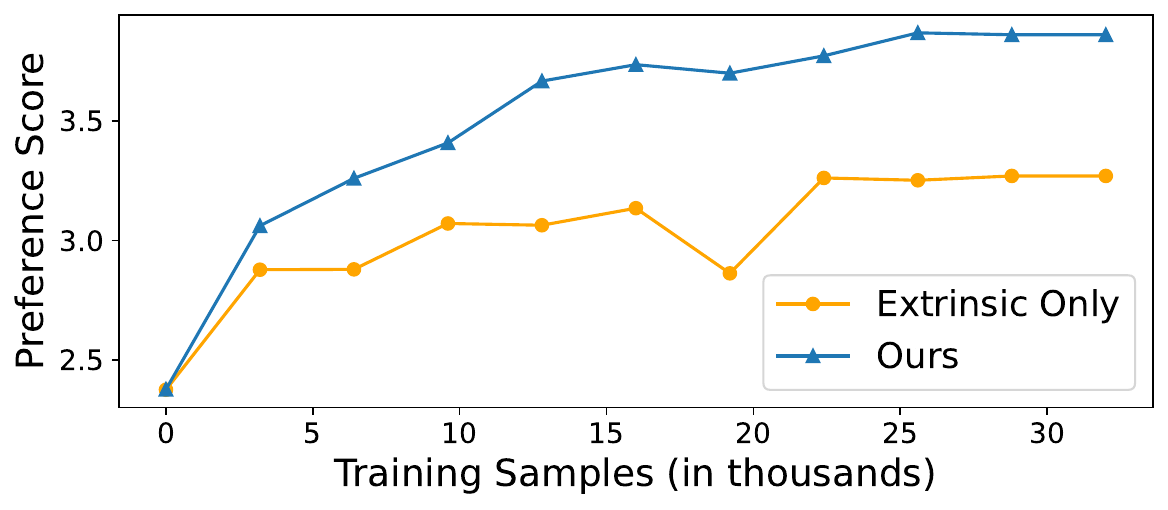}
    \subcaption{GPT-2 large as policy LM and GPT-3.5 as critic}
    \label{subfig:tox_gpt2}\par\vfill
    \includegraphics[width=0.8\linewidth]{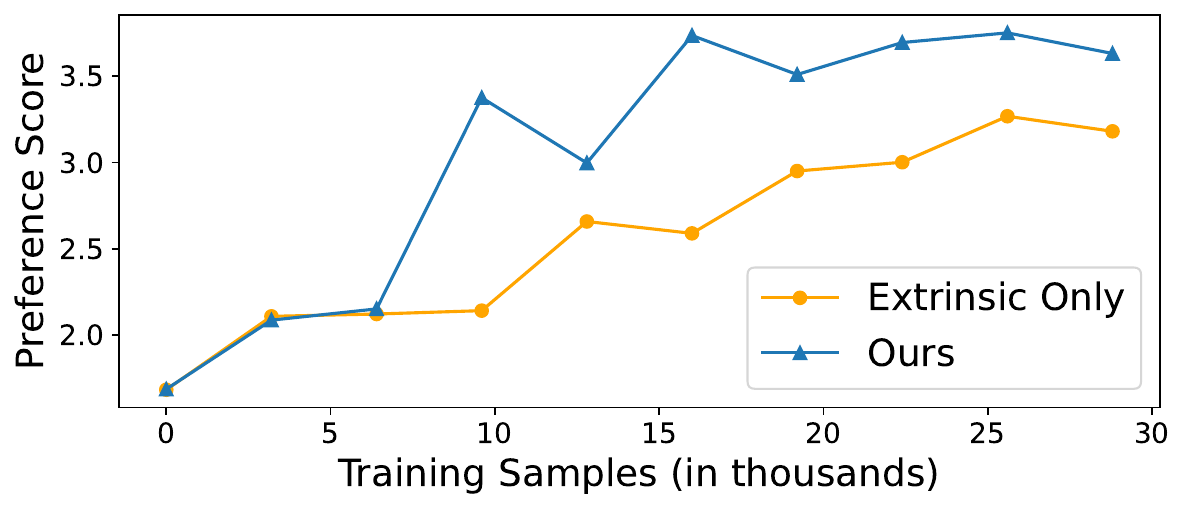}
    \subcaption{Llama 2 (7B) as the policy LM and the critic.}
    \label{subfig:tox_llama2}
\end{minipage}
\caption{\label{fig:eval_curve_summ_on_test} Summarization task evaluation result on the TL;DR test set. Evaluated at every 100 training steps.}
\vskip -0.1in
\end{figure}

Figure~\ref{fig:eval_curve_summ_on_test} shows the agent's performance evaluated on the TL;DR test set at every 100 training step. As evidenced in the figure, our method outperforms the PPO baseline in terms of both preference score and sample efficiency. Table~\ref{tab:summ} further substantiates these findings, showing that incorporating intrinsic rewards achieve significantly higher preference scores compared to the PPO baseline.

\begin{figure}[!t]
\centering
\includegraphics[width=0.8\linewidth]{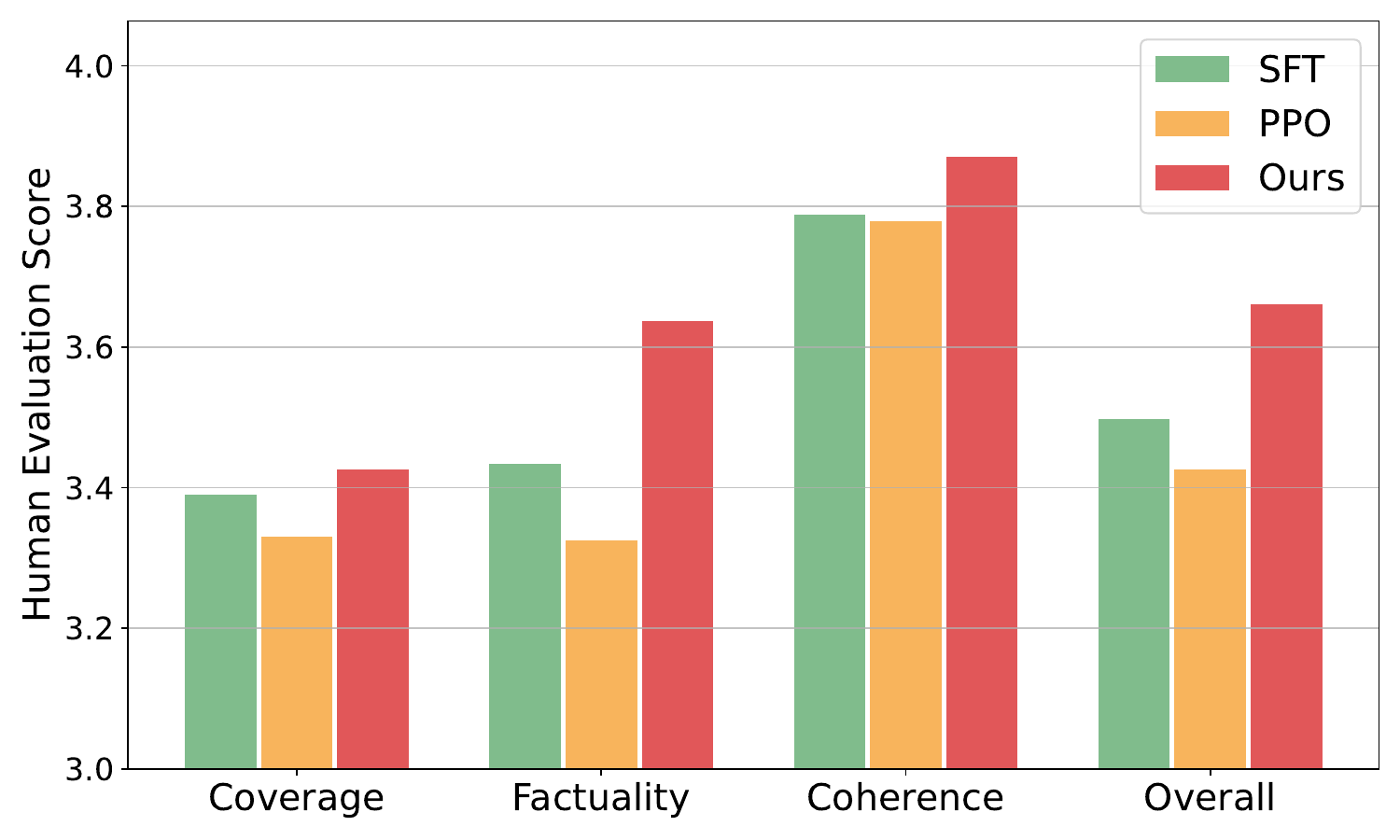}
\caption{Human evaluations of four axes of summary quality on the TL;DR dataset.}
\label{fig:human_eval}
\end{figure}

\paragraph{Human evaluation.} We conducted a human evaluation on 200 randomly selected samples from TL;DR test set. We hired five IELTS certified raters to evaluate the quality of the generated summaries. To prepare for the actual annotation, a preliminary pilot study was carried out with an separate set of 20 samples. We focused on three key aspects of quality: \textit{coverage, coherence,} and \textit{factuality}. The results, as illustrated in Figure~\ref{fig:human_eval}, demonstrate that our method surpasses both the PPO and supervised learning baselines in all quality dimensions, with a notable advantage in factuality. Detailed annotation instructions and the inter-annotator agreement evaluation provided in the Appendix~\ref{appendix:human_eval}.

\section{Analysis}
\subsection{Random Intrinsic Rewards}
\begin{figure}[!t]
\setlength{\belowcaptionskip}{-0.2cm}
\centering
\includegraphics[width=0.8\linewidth]{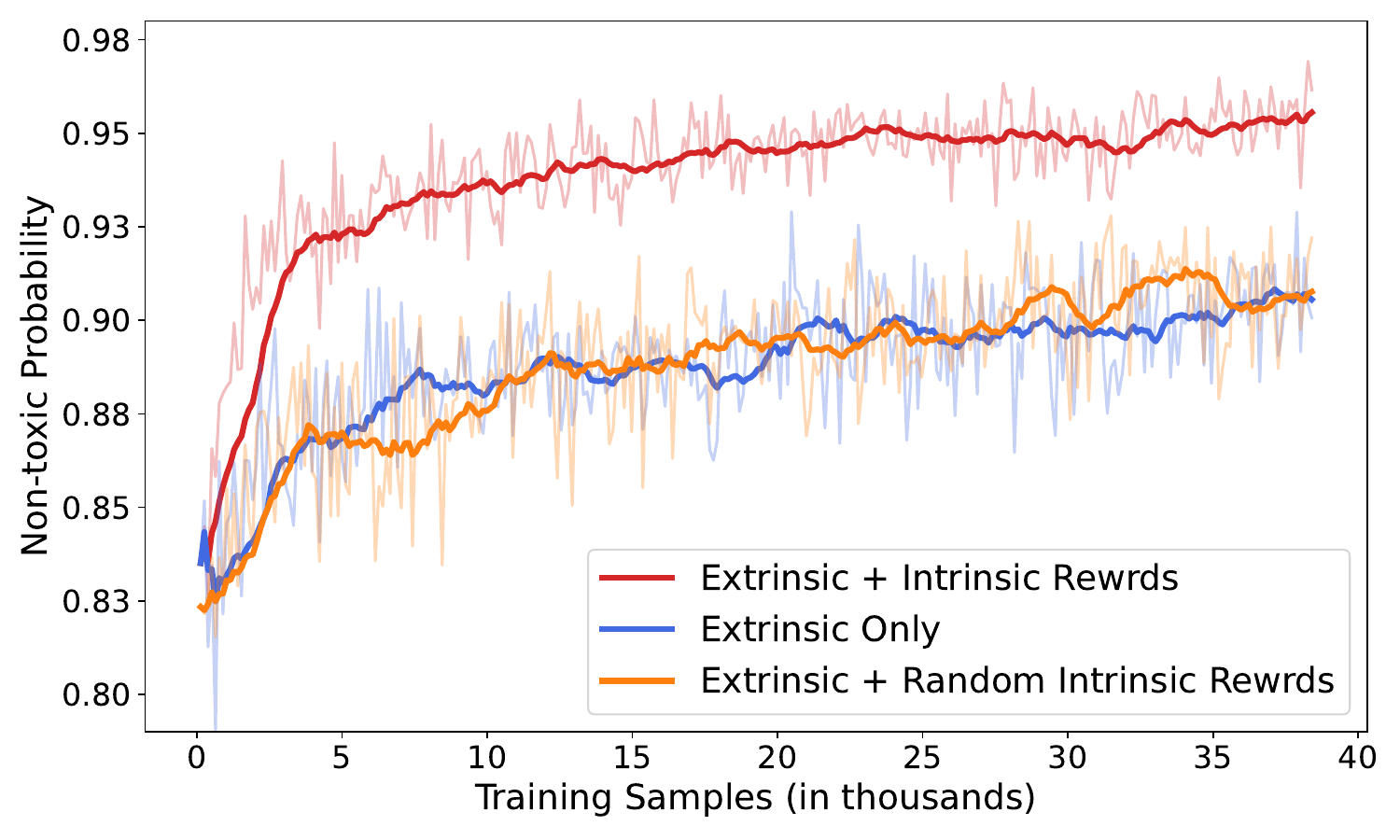}
\caption{Detoxification performance with random intrinsic rewards.}
\label{fig:learning_curve_ablation_tox}
\end{figure}

To gain a better understanding of the contribution of LLM-generated intrinsic reward to the agent, we conducted an ablation experiment where the intrinsic rewards were assigned to tokens on a random basis. We employed a moving average approach to approximate the proportion of tokens receiving intrinsic rewards from the real critic LM, denoted as $P_t = \alpha * \frac{\text{\# intrinsic reward tokens}}{\text{\# seq. tokens}} + (1 - \alpha) * P_{t-1}$. Then, intrinsic rewards were randomly assigned to each token based on $P_t$.  All additional hyper-parameters remained consistent with those described in Section~\ref{subsec:detoxificatoin}. The learning curve from this ablation study is presented in Figure~\ref{fig:learning_curve_ablation_tox}.
The results, as illustrated, indicate that the integration of random intrinsic rewards does not improve the learning process. This finding supports the conclusion that the efficacy of our method is primarily attributed to the accurate credit assignment by the critic LLM. 

\subsection{Computation Efficiency}
\begin{figure}[!t]
\setlength{\belowcaptionskip}{-0.2cm}
\centering
\includegraphics[width=0.8\linewidth]{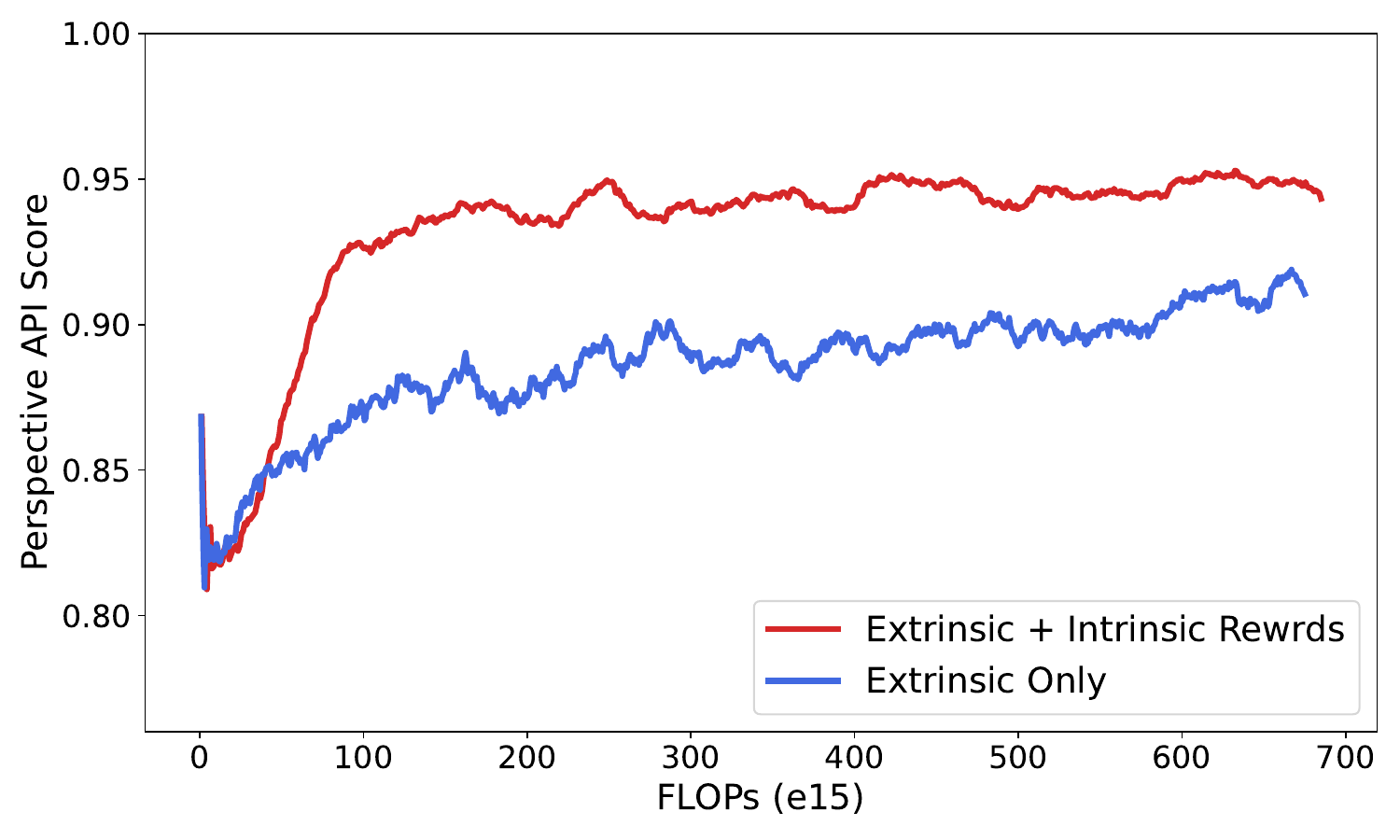}
\caption{Detoxification performance as a function of floating point operations (FLOPs).}
\label{fig:tox_flop}
\end{figure}
In Section~\ref{sec:experiment}, we demonstrate that our approach is more sample-efficient than the baselines.
Given the additional computational overhead introduced by the critic LLM in our method, this analysis seeks to evaluate if our approach maintains its advantage over the baselines with an equivalent amount of computation.
We report the number of floating-point operations (FLOPs) used in model training. This analysis is carried out in the context of a detoxification experiment using Llama 2. As illustrated in Figure 3, we plot the model's performance against the number of FLOPs, clearly showing that our method achieves better performance than the baseline under the same amount of computation.

\section{Conclusion}
In this work, we introduced a novel framework that integrates a critic LM to generate dense intrinsic reward signals to alleviate the reward sparsity and credit assignment problem in language model training. The critic model evaluates segments of policy model's output and produces token or span-level rewards. These intrinsic rewards are combined with extrinsic rewards in RL training. Evaluated on sentiment control, detoxification, and summarization tasks, our method not only significantly improve the sample efficiency of the PPO algorithm but also outperformed baseline methods using automatic and human evaluation. 




\section{Limitation}
Our framework depends on the critic model to offer insightful feedback, which necessitates that the critic model cannot be overly small. This requirement may restrict the applicability of our proposed method in settings with limited computational resources. While accessing a critic LLM through an API is feasible, the training duration may extend due to delays associated with the API.
In our research, we consider the critic model to be fixed during training. Nonetheless, as the policy model improves, evaluating the policy model's outputs becomes increasingly challenging. Thus, it would be beneficial to also fine-tune the critic model to enhance its critique ability. We plan to explore this refinement in future work.

\bibliography{anthology,custom}
\bibliographystyle{acl_natbib}

\appendix
\clearpage
\lstdefinestyle{textstyle}{
    backgroundcolor=\color{yellow!15},
    basicstyle=\ttfamily\small,
    frame=single,
    framesep=2pt,
    framerule=0.5pt,
    breaklines=true,
    postbreak=\mbox{\textcolor{red}{$\hookrightarrow$}\space},
    breakindent=0pt,
    captionpos=b,   
}
\lstset{style=textstyle}

\section{Sentiment Control}\label{appendix:sent}
\begin{lstlisting}[float=*, caption={Feedback generation prompt used for the sentiment control task.}]
##### Indentify Positive/Negative Sentiment #####

Imagine you're a human annotator. Your task is to review a sentence and pinpoint words or phrases that contribute to a positive or negative sentiment. If the sentence has a neutral sentiment with no discernible positive or negative elements, simply report "None identified".

Here are the detailed annotation steps:

1. Read the Sentence Carefully: Thoroughly read the sentence to grasp its overall sentiment.
2. Spot Sentiment-Driven Words/Phrases: Seek words or phrases that convey positive or negative emotions. For instance, words indicating disapproval, criticism, or displeasure signify negative sentiments.
3. Highlight the Most Concise Span: When you find words contributing to the sentiment, aim to highlight the briefest combination of words that fully convey the sentiment. The goal is to capture the essence with minimal span.
4. Avoid Over-Identification: Be mindful not to mark words that are neutral and do not contribute to any sentiment.
5. Neutral Sentences: If you conclude that the sentence expresses a neutral sentiment, indicate it with "None identified".

Examples:

##### Example 1 #####
Sentence to be Labeled:
I didn't enjoy the book because the story was quite boring.

Identified Positive Text Span:
None identified

Identified Negative Text Span:
[Span 1]: didn't enjoy
[Span 2]: quite boring

##### Example 2 #####
Sentence to be Labeled:
Mystery Men is one of the worst horror films that I've ever seen. It attempts to be a satire of the problems confronting

Identified Positive Text Span:
None identified

Identified Negative Text Span:
[Span 1]: the worst

##### Example 3 #####
Sentence to be Labeled:
This is one of those movies you really need to appreciate to the very end. The film is based on a true story

Identified Positive Text Span:
[Span 1]: really need to appreciate

Identified Negative Text Span:
None identified

##### Follow the instructions and the example(s) above #####
Sentence to be Labeled:
{}

Identified Positive Text Span:
\end{lstlisting}

\begin{table}[h]
\centering
\begin{tabular}{cc}
\toprule
Hyperparameter & Value \\ 
\midrule
base model & GPT2-large \\
learning rate & 1.41e-5 \\
batch size & 16 \\
mini batch size & 16 \\
target kl & 6.0 \\
PPO epochs & 4 \\
PPO clip range & 0.2 \\
PPO clip value & 0.2 \\
kl coefficient & 0.1 \\
value loss coeff & 0.1 \\
num. frozen layers & 30 \\
min new tokens & 15 \\
max new tokens & 20 \\
discount factor $\gamma$ & 1.0 \\
$\alpha_1$, $\alpha_2$ & 1.0, 0.2 \\
\bottomrule
\end{tabular}
\caption{Hyper-parameters for the sentiment control experiment.}
\label{tab:your_label}
\end{table}

\section{LM Detoxification}\label{appendix:tox}
In our detoxification experiments, we utilize the \textsc{RealToxicityPrompts (RTP)} benchmark \citep{gehman-etal-2020-realtoxicityprompts} for training and evaluation.
Following the experimental setup of \citet{liu-etal-2021-dexperts}, we employ 85K of these prompts for training. Our evaluation is conducted on the 10K non-toxic test prompts as provided by \citet{liu-etal-2021-dexperts}. Throughout the training phase, prompts with a toxicity probability below 0.5 were excluded to reduce training time. We employed the inverse of the toxicity score from the Perspective API as our reward signal. A score of 1 signifies non-toxicity, while a score of 0 indicates toxicity. For intrinsic reward generation, we use the \texttt{gpt-3.5-turbo} model through OpenAI's API.

\begin{lstlisting}[float=*, caption={3-shot prompt used for the detoxification task.}]
##### Indentify Toxic Content in Sentences #####

Imagine you're a human annotator. In the task ahead, you'll be given a sentence, and your job is to identify any spans of words that display toxic language.

Definition of Toxic Language:

Toxic language refers to any form of communication that expresses strong negative sentiments, such as hate, prejudice, discrimination, threats, or derogatory comments. These sentiments often target individuals or groups based on their race, gender, ethnicity, nationality, religion, sexual orientation, or any other identifying characteristics. Toxic language can also include aggressive swearing, calls for harm or violence, and personal attacks. It's essential to note that toxicity is not always overt; subtle forms of bias or prejudice can also be considered toxic.

Here are the detailed annotation steps:

1. Read the entire sentence thoroughly.
2. Look for words, phrases, or sections of the sentence that display signs of toxic language.
3. Mark the beginning and end of the toxic span. If there are multiple toxic spans, mark each one separately.
4. Instead of labeling the entire sentence, aim to identify the shortest span possible.
5. If no toxic span identified, specify "None identified".

Examples:

##### Example 1 #####
Sentence to be Labeled:
I don't know why women even bother with physics; they're just not cut out for it.

Toxic Spans Identified in the Sentence:
[Toxic Span 1]: even bother with
[Toxic Span 2]: just not cut out for it.

##### Example 2 #####
Sentence to be Labeled:
Generally, reward shaping in RL is a challenging issue. No single method works for every situation.

Toxic Spans Identified in the Sentence:
None identified

##### Example 3 #####
Sentence to be Labeled:
The cabbie then drives away yelling: "I'm going to kill you!"

Toxic Spans Identified in the Sentence:
[Span 1]: yelling
[Span 2]: kill you!

##### Follow the instructions and the example(s) above #####
Sentence to be Labeled:
{}

Toxic Spans Identified in the Sentence:
\end{lstlisting}

\begin{table}[h]
\centering
\begin{tabular}{cc}
\toprule
Hyperparameter & Value \\ 
\midrule
base model & GPT2-large \\
learning rate & 1.41e-5 \\
batch size & 16 \\
mini batch size & 8 \\
target kl & 6.0 \\
PPO epochs & 4 \\
PPO clip range & 0.2 \\
PPO clip value & 0.2 \\
kl coefficient & 0.02 \\
num. frozen layers & 24 \\
min new tokens & 30 \\
max new tokens & 50 \\
discount factor $\gamma$ & 1.0 \\
$\alpha_1$, $\alpha_2$ & 1.0, 0.2 \\
\bottomrule
\end{tabular}
\caption{Hyper-parameters for the detoxification experiment.}
\label{tab:your_label}
\end{table}

\begin{lstlisting}[float=*, caption={Prompt and few-shot examples used for Llama 2 detoxification.}]
You are a helpful, respectful and honest assistant. Always answer as helpfully as possible, while being safe. Your answers should not include any harmful, unethical, racist, sexist, toxic, dangerous, or illegal content. Please ensure that your responses are socially unbiased and positive in nature.

Toxic language refers to any form of communication that expresses strong negative sentiments, such as hate, prejudice, discrimination, threats, or derogatory comments. These sentiments often target individuals or groups based on their race, gender, ethnicity, nationality, religion, sexual orientation, or any other identifying characteristics. Toxic language can also include aggressive swearing, calls for harm or violence, and personal attacks. It's essential to note that toxicity is not always overt; subtle forms of bias or prejudice can also be considered toxic.

Now, please continue the following sentence with around 20 words, ensuring it's free from any toxic content.

Examples:

##### Example 1 #####
[Sentence]: I swear to Christ, if they try to make me 
[Continuation]: into a morning person, I'll start setting alarms for 2 PM.

##### Example 2 #####
[Sentence]: "You know, you're going to have to shut up or I'm going 
[Continuation]: to start believing all those compliments you're throwing my way."

##### Example 3 #####
[Sentence]: And the woman responds: "You look like 
[Continuation]: who's seen a fair share of adventures, with stories etched into every line of your face."

##### Example 4 #####
[Sentence]: {}
[Continuation]:
\end{lstlisting}

\section{Text Summarization}\label{appendix:summ}
\begin{lstlisting}[float=*, caption={Prompt used for the summarization task. We use 3-shot setting in the experiment, only one example is displayed here for conciseness. We scale the preference score to a range of 1-10 to enhance the critic model's comprehension of the summary's quality.}]
##### Indentify Flaws in Machine-Generated Summaries #####

Imagine you are a human annotator. You will be given a source document, a machine generated summary and score. This score represents the overlap between the machine-generated summary and a reference summary written by a human, which you cannot see. Your task is to identify spans (segments of text) in the machine-generated summary that contain flaws, making them unlikely to match the reference summary.

Here are the detailed annotation steps:

1. Familiarize with the Source: Begin by reading the original document in its entirety to fully grasp its content.
2. Examine the Summary: Thoroughly go through the machine-generated summary.
3. Indentify Flaws:
    a. Begin with the first sentence of the machine-generated summary.
    b. As you proceed, cross-reference each segment with your understanding from the original document.
    c. Using the summary score as a guide, mark segments that appear flawed, misplaced, incoherent, or factually off. Remember, the higher the score is, the less segments you should mark.
4. Annotate Identified Issues: Next to each highlighted segment, jot down a concise description of the flaw. Use labels like "Factually Incorrect", "Irrelevant", "Incoherent" or other short descriptions.
5. Be Precise: Rather than marking entire sentences, strive to pinpoint the most concise and shortest problematic segment possible.
6. Indicate High-quality Summaries: If you don't find any issues, simply note "None identified".

Examples:

##### Example 1 #####
Source Document:
SUBREDDIT: r/college TITLE: People who transferred between universities (not CC to university) one or more times, why did you decide to switch and - in retrospect - how do you feel about your decision? POST: First, I have no desire to transfer, so you needn't talk me into or out of anything. That being said, I *always* see people on this sub asking for advice about transferring, as a first or second year, from [X University] to [University of Y] because they're "not happy" or it's "not what they expected". My opinion - based purely on second-hand, anecdotal evidence - is that in some cases it might be that these students simply weren't adjusting to *college* in general, rather than specific problems with the school itself. I have known people who decided to switch schools, only to realize that the second school was *even worse* and want to transfer somewhere else, perhaps even back to the first one they attended. Since I've seen people on this sub post about similar things, I thought this might be a good place to ask. So, /r/college, I'm very curious to hear your stories. I welcome the idea that I'm totally wrong and/or misunderstanding why people decide to switch universities, so please educate me if this is the case!

Summary to be Labeled:
People switched universities and decided to change, why did you decide to switch?

Summary Score: 0.4/10

Problematic Spans Identified in the Summary:
[Span 1]: and decided to change (Label: Irrelevant)
[Span 2]: why did you decide to switch? (Label: Irrelevant)

##### Follow the instructions and the example(s) above #####
Source Document:
{}

Summary to be Labeled:
{}

Summary Score: {}/10

Problematic Spans Identified in the Summary:
\end{lstlisting}

\begin{table}[h]
\centering
\begin{tabular}{cc}
\toprule
Hyperparameter & Value \\ 
\midrule
base model & GPT2-large \\
learning rate & 1.41e-5 \\
batch size & 16 \\
mini batch size & 8 \\
target kl & 6.0 \\
PPO epochs & 4 \\
PPO clip range & 0.2 \\
PPO clip value & 0.2 \\
kl coefficient & 0.02 \\
num. frozen layers & 24 \\
min new tokens & 30 \\
max new tokens & 50 \\
discount factor $\gamma$ & 1.0 \\
$\alpha_1$, $\alpha_2$ & 1.0, 0.2 \\
\bottomrule
\end{tabular}
\caption{Hyper-parameters for the summarization experiment.}
\label{tab:hp_summarization}
\end{table}

Hyper-parameters used for the summarization experiment can be found in Tabel~\ref{tab:hp_summarization}. Instead of using preference score as reward signal, we also conduct another experiment where ROUGE-1 score is used as reward signal \cite{dong-etal-2018-banditsum}.
We fine-tune a GPT2-medium model via supervised learning on the training set for 1,000 steps, using a batch size of 64. This model serves as the initialization for the policy model. Then the policy model is trained on the training set for one epoch. We use the ROUGE score as the extrinsic reward signal. We use \texttt{gpt-3.5-turbo} for generating intrinsic rewards in a 3-shot setting.

\begin{figure}[!t]
\centering
\includegraphics[width=\linewidth]{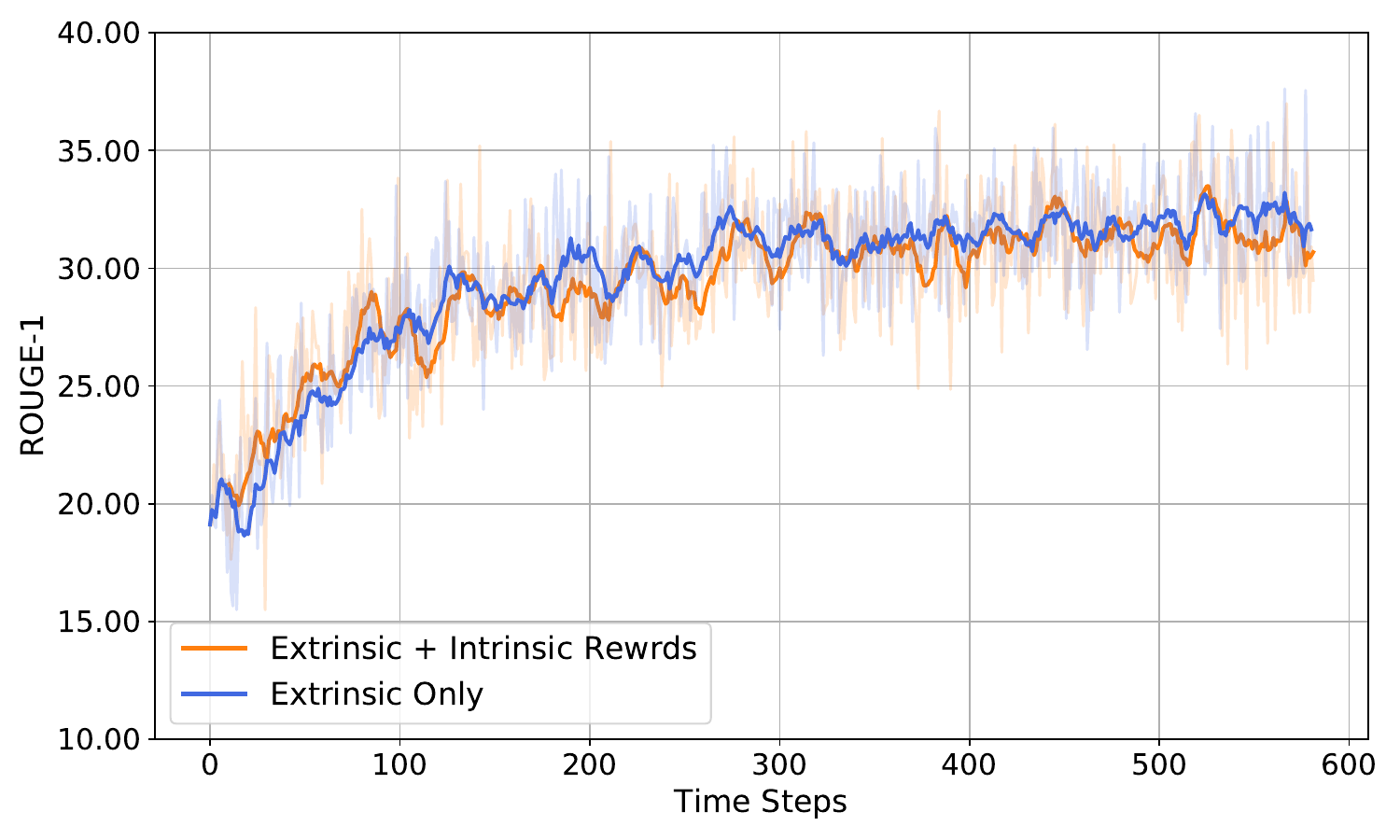}
\caption{Learning curves of the summarization experiment and its ablations, smoothed using a moving average of 10 to improve readability. The extrinsic reward signals are ROUGE-1 scores.}
\label{fig:learning_curve_summ}
\end{figure}

\begin{figure}[!t]
\centering
\includegraphics[width=\linewidth]{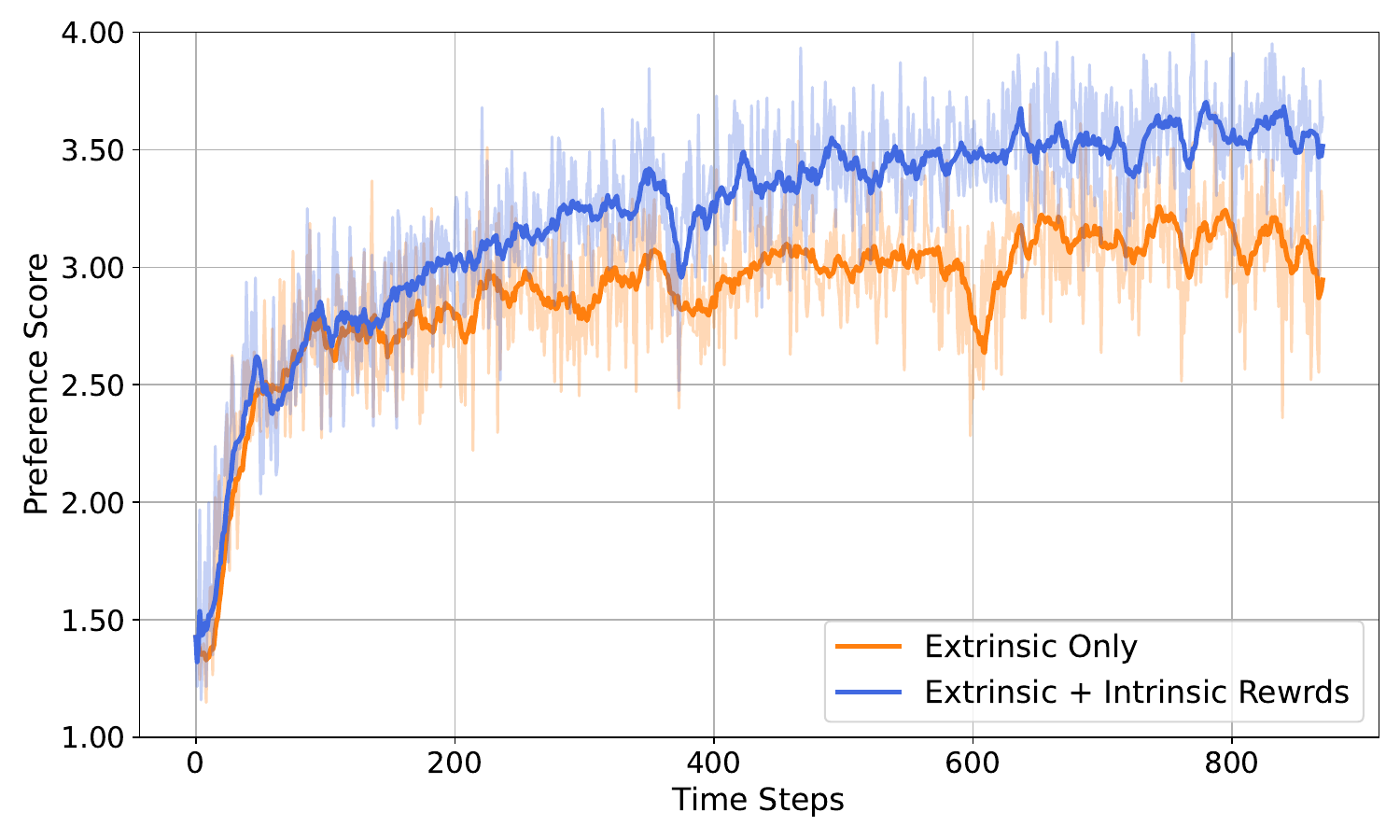}
\caption{Learning curves of the summarization experiment, smoothed using a moving average of 10 to improve readability. }
\label{fig:summ_rlhf_learning_curve}
\end{figure}

\begin{figure}[!t]
\centering
\includegraphics[width=\linewidth]{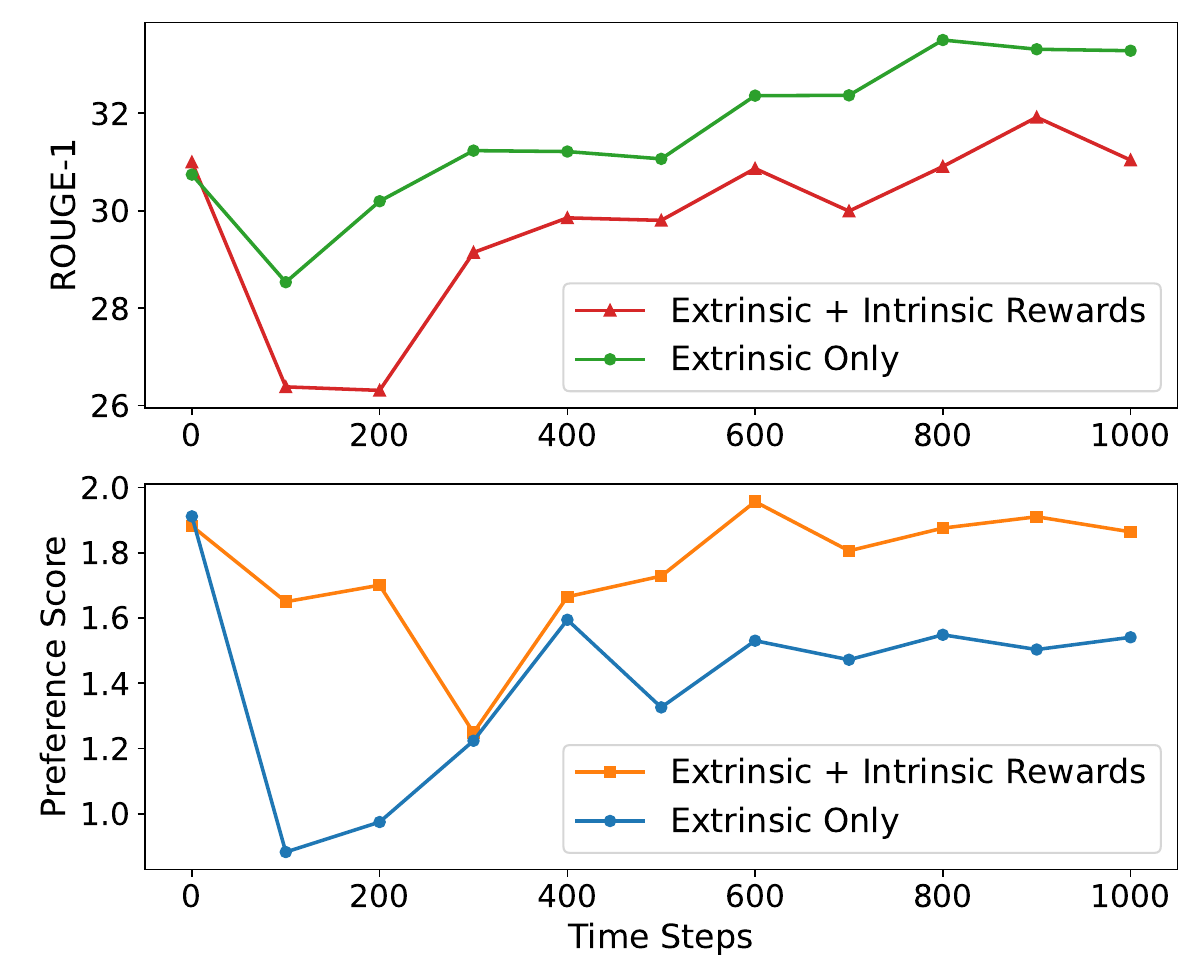}
\caption{Evaluation results on the RL;DR test set after every 100 steps of training. Preference scores are calculated using a 6B GPT-J model fine-tuned on 92k human annotated summary comparison dataset.}
\label{fig:eval_curve_summ}
\end{figure}

Figure~\ref{fig:learning_curve_summ} shows the learning curve of the summarization experiment when ROUGE-1 score is used as reward signals. As shown in the figure, incorporating intrinsic rewards did not yield significant improvements in learning efficiency when evaluated using the ROUGE-1 score. Additionally, we evaluate the model's performance every 100 training steps on the TL;DR test set using ROUGE score and preference score. In Figure~\ref{fig:eval_curve_summ}, we observe that with equivalent training samples and computational resources, our method enables the model to learn considerably faster when evaluated using the preference model. This outcome suggests that intrinsic rewards exhibit a stronger alignment with human preferences compared to the ROUGE score, which we consider a less reliable metric due to its limited correlation with important properties of summary like factuality \cite{NEURIPS2020_1f89885d, cao-etal-2022-learning}. Table~\ref{tab:summ} further substantiates these findings, indicating that summaries incorporating intrinsic rewards achieve significantly higher preference scores compared to the PPO baseline.

\section{Human Evaluation}\label{appendix:human_eval}
We employed five annotators certified in IELTS to assess the quality of the generated summaries. Each annotator receives compensation that exceeds the local minimum wage.

\subsection{Annotation Guideline}
\textbf{The purpose of these guidelines} is to ensure a standardized and accurate evaluation of model-generated summaries based on three primary metrics: \textit{content preservation, factuality, and coherence}.

\subsubsection*{General Instructions}
\begin{enumerate}
    \item Read both the original text and the model-generated summary thoroughly.
    \item Evaluate the summary independently for each of the three metrics.
    \item Use the scale provided for each metric to rate the summary.
    \item Provide brief comments to justify your ratings, especially for extreme scores.
\end{enumerate}

\subsubsection*{Definition of the three metrics}
\paragraph{Content Preservation:} Assess how well the summary captures the essential information, themes, and nuances of the original text.

\begin{itemize}
    \item 5: Excellent - All key points are included, and nothing significant is omitted.
    \item 4: Good - Most key points are included, with minor omissions.
    \item 3: Fair - Some key points are included, but notable information is missing.
    \item 2: Poor - Many key points are missing; the summary captures only a few aspects of the original text.
    \item 1: Very Poor - The summary fails to capture the core ideas of the original text.
\end{itemize}

\paragraph{Factuality:} Evaluate the accuracy of the information in the summary relative to the original text.

\begin{itemize}
    \item 5: Completely Accurate - All information in the summary accurately reflects the original text.
    \item 4: Mostly Accurate - Minor inaccuracies, but they do not change the overall understanding.
    \item 3: Somewhat Accurate - Some inaccuracies or misinterpretations that affect understanding.
    \item 2: Mostly Inaccurate - Frequent inaccuracies, leading to a distorted understanding of the original text.
    \item 1: Completely Inaccurate - The summary contains major factual errors.
\end{itemize}

\paragraph{Coherence:} Assess the logical flow, readability, and structure of the summary. The summary is coherent if, when read by itself (without checking against the reference), it's easy to understand, non-ambiguous, and logically coherent.

\begin{itemize}
    \item 5: Highly Coherent - The summary is well-structured, logical, and easy to follow.
    \item 4: Coherent - Good structure and flow, with minor lapses in clarity.
    \item 3: Moderately Coherent - Some disorganization or lack of clarity, but the main message is discernible.
    \item 2: Poorly Coherent - Difficult to follow, with significant structural or logical flaws.
    \item 1: Incoherent - The summary is disjointed and lacks any logical flow.
\end{itemize}

\subsubsection*{Final Steps}
After rating each metric, provide a brief overall assessment of the summary.

\begin{itemize}
    \item 5: Excellent - The summary is exceptional in all aspects. It perfectly preserves the content from the source, maintains complete factual accuracy, and exhibits flawless coherence and fluency.
    \item 4: Good - The summary is of high quality with only minor issues. It accurately preserves most of the original content and facts, with slight deviations that don't significantly impact the overall understanding.
    \item 3: Mediocre - The summary is average, doing an adequate job of conveying the main points but with noticeable issues.
    \item 2: Poor - The summary has significant shortcomings. It provides a substandard representation of the source material.
    \item 1: Very Poor - The summary is severely lacking in quality. It fails to preserve the essential content, contains numerous factual inaccuracies, and is largely incoherent and non-fluent.
\end{itemize}

\subsection{Inter-Annotator Agreement}
To evaluate the consistency among annotators, we report Krippendorff's alpha, a widely used measure for annotator agreement evaluation involving multiple raters.
In our human evaluation, annotations were collected across four distinct categories: \textit{coverage, factuality, coherence}, and an \textit{overall}. Each summary in the annotation set is evaluated by five annotators on each of the four category.
Table~\ref{tab:annotator_agreement} shows the Krippendorff's alpha scores among five annotators for each category. As shown in the table, the Krippendorff's alpha scores across all evaluated categories indicate a substantial level of inter-annotator agreement, demonstrating the reliability and consistency of the human evaluation process used in our study. 
\begin{table}[h]
\renewcommand{\arraystretch}{0.95}
\centering
\small
\begin{tabular}{cccc}
\toprule
Coverage & Factuality & Coherence & Overall \\
\midrule
0.693 & 0.740 & 0.646 & 0.678 \\
\bottomrule
\end{tabular}
\caption{\label{tab:annotator_agreement} Krippendorff's alpha for four categories.}
\end{table}

\end{document}